\documentclass[letterpaper, 10 pt, conference]{ieeeconf}  
\IEEEoverridecommandlockouts                            
\usepackage{amssymb}
\usepackage{pifont}             
\usepackage{graphicx}
\usepackage{floatrow}
\usepackage{makecell}
\newfloatcommand{capbtabbox}{table}[][\FBwidth]
\setcellgapes{3pt}

\usepackage{amsmath}
\usepackage[colorinlistoftodos]{todonotes}

\usepackage{wrapfig,lipsum,booktabs}

\definecolor{MyDarkBlue}{rgb}{0,0.08,1}
\definecolor{MyDarkGreen}{rgb}{0.02,0.6,0.02}
\definecolor{MyDarkRed}{rgb}{0.8,0.02,0.02}
\definecolor{MyDarkOrange}{rgb}{0.40,0.2,0.02}
\definecolor{MyPurple}{RGB}{111,0,255}
\definecolor{MyRed}{rgb}{1.0,0.0,0.0}
\definecolor{MyGold}{rgb}{0.75,0.6,0.12}
\definecolor{MyDarkgray}{rgb}{0.66, 0.66, 0.66}
\definecolor{MyDarkBlack}{rgb}{0 0 0}

\usepackage[footnotesize]{caption}
\usepackage{color,soul}
\makeatletter
\let\NAT@parse\undefined
\makeatother
\usepackage[numbers,sort&compress]{natbib}
\usepackage{multirow}

\title{A Hierarchical Structure for Dexterous Manipulation: When Learning Meets Dynamic Control}
\title{A Hierarchy for Dexterous Manipulation: \\When Learning meets Dynamic Control}
\title{Learning meets Control for In-Hand Manipulation}
\title{Learning meets Control for Dexterous Manipulation}
\title{Hierarchical Control Learning for Robust Dexterous Manipulation}
\title{Learning Hierarchical Control for Robust Dexterous Manipulation}
\title{Hierarchical Control Policies for Robust Dexterous Manipulation}
\title{Hierarchical Policy Learning for Dexterous Manipulation}
\title{A Hierarchy for Dexterous Manipulation: \\When Learning meets Control}
\title{Learning Complex Dexterous Manipulation}
\title{Learning Robust Hierarchical Control for Dexterous Manipulation}
\title{Learning Hierarchical Control for Dexterous Manipulation}
\title{Learning Hierarchical Control\\for In-Hand Manipulation}
\title{Learning Robust Hierarchical Control\\for In-Hand Manipulation}
\title{\bf \LARGE Learning Hierarchical Control for Robust In-Hand Manipulation}

\author{Tingguang Li$^{1,2}$, Krishnan Srinivasan$^{2}$, Max Qing-Hu Meng$^{1}$, Wenzhen Yuan$^{3}$ and Jeannette Bohg$^{2}$
\thanks{$^{1}$T. Li and M. Meng are with the Department of Electronic Engineering, The Chinese University of Hong Kong, Hong Kong, China. \{\tt\small tgli, Max\}@ee.cuhk.edu.hk}%
\thanks{$^{2}$T. Li, K. Srinivasan and J. Bohg are with the Computer Science Department, Stanford University, USA. {\tt\small \{tgli, krshna, bohg\}@stanford.edu}}%
\thanks{$^{3}$W. Yuan is with the Robotics Institute, Carnegie Mellon University, USA. {\tt\small yuanwz@cmu.edu}}%
}

\begin{document}
\maketitle
\thispagestyle{empty}
\pagestyle{empty}

\begin{abstract}
Robotic in-hand manipulation has been a long-standing challenge due to the complexity of modelling hand and object in contact and of coordinating finger motion for complex manipulation sequences. To address these challenges, the majority of prior work has either focused on model-based, low-level controllers or on model-free deep reinforcement learning that each have their own limitations. We propose a hierarchical method that relies on traditional, model-based controllers on the low-level and learned policies on the mid-level. The low-level controllers can robustly execute different manipulation primitives (reposing, sliding, flipping). The mid-level policy orchestrates these primitives. We extensively evaluate our approach in simulation with a 3-fingered hand that controls three degrees of freedom of elongated objects. We show that our approach can move objects between almost all the possible poses in the workspace while keeping them firmly grasped. We also show that our approach is robust to inaccuracies in the object models and to observation noise. Finally, we show how our approach generalizes to objects of other shapes.
\end{abstract}

\section{Introduction}
{\em Dexterous Manipulation\/} refers to the ability of changing the pose of an object to any other pose within the workspace of a hand~\cite{okamura2000overview,Bicchi2000,ma2011dexterity}. In this paper, we are particularly concerned with the ability of in-hand manipulation where the object is continuously moved within the hand without dropping. This ability is used frequently in human manipulation e.g. when grasping a tool and re-adjusting it within the hand, when inspecting an object, when assembling objects or when adjusting an unstable grasp. Yet, in-hand manipulation remains a long-standing challenge in robotics despite the availability of multi-fingered dexterous hands such as~\cite{MITUtahHand,GifuHand,AllegroHand}.

\begin{figure}[]
    \centering
    \small
    \includegraphics[width=1\textwidth]{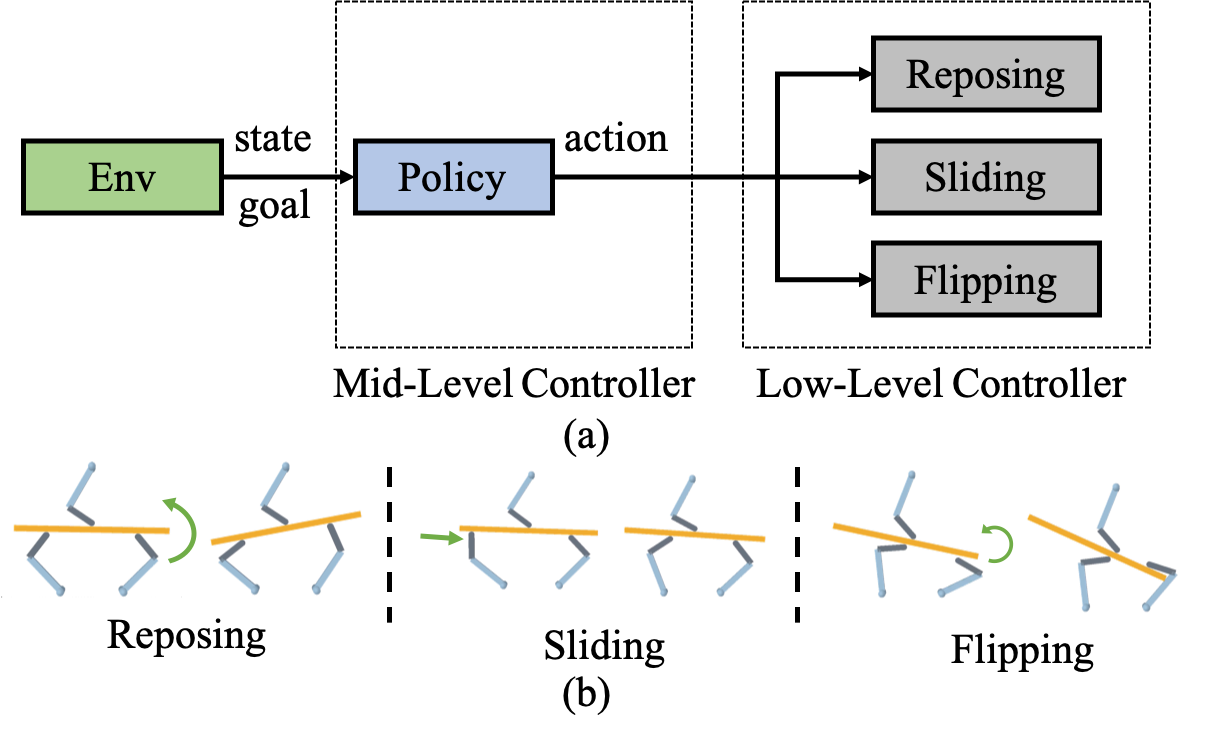}
    \vspace{-18pt}
    \caption{
    (a) The proposed two-level hierarchical architecture for in-hand manipulation. At the low level, we have three manipulation primitives that are executed with low-level, torque controllers. At the mid-level, we train a policy with DRL to pick a manipulation primitive and its parameterization. (b) Visualization of the three manipulation primitives. 
    }
    \label{fig:architecture}
\end{figure}
Controlling dexterous manipulation can be divided into three levels~\cite{okamura2000overview}: (i) High-level control for task and motion planning as well as grasp selection, (ii) mid-level control for smoothly coordinating and sequencing manipulation phases towards a goal pose and (iii) low-level control that tracks the reference signal provided by the mid-level control with e.g. force or impedance controllers. The low-level has received considerable attention in prior work~\cite{rus1992dexterous,okamura2000overview,Bicchi2000,li2013integrating,rus1997coordinated,sundaralingam2017relaxed}. These methods assume some prior knowledge of the geometry and inertia of hand and object as well as of contact locations. They can robustly control local adjustments of the object pose in-hand. However, mid-level control is required to sequence manipulation phases and thereby to reach more distant object poses. Compared to low and high-level control, mid-level control is not as well studied \cite{okamura2000overview, calli2016vision}.

More recently, {\em Deep Reinforcement Learning\/} (DRL) has been used to learn dexterous manipulation. These are typically end-to-end approaches that do not assume a division between the low- and mid-level \cite{kumar2016optimal,OpenAIHand,Zhu2018DexterousMW}. While these approaches do not require prior knowledge of hand and object properties, policies are learned per object instance. Also, these works are typically demonstrated on manipulation tasks that do not require the object to be continuously held by the hand. For example, the palm may always be facing upwards to prevent the object from dropping~\cite{OpenAIHand,kumar2016optimal}, or the hand is manipulating an articulated object that is rigidly attached in the world~\cite{Zhu2018DexterousMW}. Learning to hold the object firmly while reposing would likely require DRL methods an even larger amount of training episodes and a very carefully designed reward function. Nevertheless, in many real-world manipulation tasks this is an important ability. 

We are proposing a hierarchical control structure for in-hand manipulation of objects within the entire reachable workspace. At the low-level, we use well-studied, compliant torque controllers per manipulation primitive. At the mid-level, we use DRL to learn a policy that sequences manipulation primitives towards a more distant object goal pose. Through this combination, we ensure that the object is continuously held by the hand. While the low-level controllers require prior knowledge about hand and object properties, we show that our framework is robust against observation noise in the object pose and inaccuracies of kinematic and dynamic models and thereby generalizes to object variations.


We define three manipulation primitives: reposing, sliding, and flipping, as shown in Fig. \ref{fig:goal_vis}. Reposing, or in-grasp manipulation \cite{sundaralingam2018geometric}, refers to moving the object while keeping the same contact configuration. Sliding and flipping change the contact point of one finger by either sliding it along the surface or moving it to a different side of the object. Different contact configurations have different ranges of poses that the object can reach. We train a policy with DRL to first reach a feasible contact configuration of the object and then select among these three manipulation primitives to repose the object to the goal pose. 
We perform an extensive evaluation of our approach in a simulated environment consisting of a three-fingered hand that can change the 3DoF pose (2D position and orientation) of pole-shaped objects. We also show that the overall hierarchical approach is suitable for other object shapes.

Compared to a graph search-based approach~\cite{cruciani2019dual} and an end-to-end baseline, our method has a significantly higher success rate of reaching a target object pose for a pole, especially when they are far from the initial poses and complex finger coordination is required.  
\section{Related Work}
\subsection{Model-based Approaches to Dexterous Manipulation}
Traditional approaches towards dexterous manipulation typically assume known kinematics, dynamics and contact models. They devise controllers that execute 
primitives such as reposing/in-grasp manipulation, finger sliding or finger gaiting. \citet{rus1992dexterous,rus1997coordinated} developed different methods of rotating an object in a plane, involving a force-direction closure grasp. 
\citet{okada1982computer,kerr1986analysis} applied sliding and rolling of robot fingers to manipulate objects of various shapes using inverse kinematics. 
A typical challenge for model-based approaches is to reach more distant object poses that requires longer manipulation sequences which switch between different primitives.
To address this and plan a sequence of primitives, early work by \cite{leveroni1996reorienting} proposed a finger gaiting scheme based on geometrical modeling. \citet{mordatch2012contact} developed a trajectory optimization method that uses a contact-invariant cost. \citet{fan2017real} developed a two-staged planner that first plans finger gaits and then uses a controller to move the object. However, neither of these methods has been demonstrated to be robust to noise and inaccurate models.
\citet{odhner2015stable,calli2018robust} propose a method to roll or slide objects in-hand. Their method uses model-predictive control within a visual-servoing scheme. The authors specifically focus on underactuated hands with a low DoF while we consider high-DOF, fully-actuated hands.
\citet{platt2004manipulation} define several conceptual states to describe the space of grasp configurations and explore the manipulation behavior in the context of a Markov Decision Process. \citet{simpkins2011complex} developed a hierarchical control scheme that can plan ideal forces and torques. \citet{cruciani2019dual,murooka2017global} formulated the manipulation problem as a graph and generated the manipulation sequence by searching for a path in that graph. These approaches typically make strong assumptions like sticky contacts with infinite friction~\cite{simpkins2011complex}, fully-known, deterministic transition models~\cite{cruciani2019dual,murooka2017global} and a finite state space~\cite{platt2004manipulation}, which make them less likely to cope with real world challenges. 

\subsection{Model-Free Approaches to Dexterous Manipulation}
\label{sec:model-free}
Different from traditional methods, model-free approaches towards dexterous manipulation use limited or no prior knowledge of the gripper or object. They learn to control the object through reinforcement learning typically without distinguishing between the low and mid-level. These end-to-end learning approaches learn a mapping between observations of the object and control commands of the robot hand.
\citet{kumar2016optimal} developed a sample-efficient method for training an anthropomorphic hand to rotate a cylinder. At the core, they optimize linear-gaussian controllers via iterative-LQR.
\citet{OpenAIHand} train object-specific DRL policies for in-hand reposing. \citet{van2015learning} train a policy that uses tactile sensory data as input. The robot learns to roll and reorient cylindrical objects on a table. The policy is robust to changes in the size and weight of the objects. 
However, neither work requires the robot to robustly hold the object while manipulating it as the object is typically supported by either the upward-facing palm of the hand, a table or it is an articulated object fixed to the environment.

Furthermore, model-free DRL has a high sample complexity. To address this, \cite{rajeswaran2017learning,Zhu2018DexterousMW} combined imitation learning and reinforcement learning.
However, the considered tasks are also constructed such that the object does not need to be grasped continuously and does not easily drop outside the reach of the hand. Learning a policy from scratch that firmly holds an object during in-hand manipulation may require excruciatingly high number of episodes and carefully tuned reward function.  


\subsection{Combining Deep Reinforcement Learning With Models}
We propose a hybrid framework to combine the advantages of both, traditional model-based and more recent model-free approaches.
To this end, \citet{silver2018residual,johannink2018residual} propose to learn residual policies that add corrective joint-commands to an existing controller. Thereby they account for model inaccuracies and noise. Our method is hierarchical in structure where the DRL policy selects manipulation primitives to be executed by the low-level controller, as opposed to being compositional, where the policy and controller output are summed.

\begin{figure}[]
    \centering
    \small
    \includegraphics[width=\textwidth]{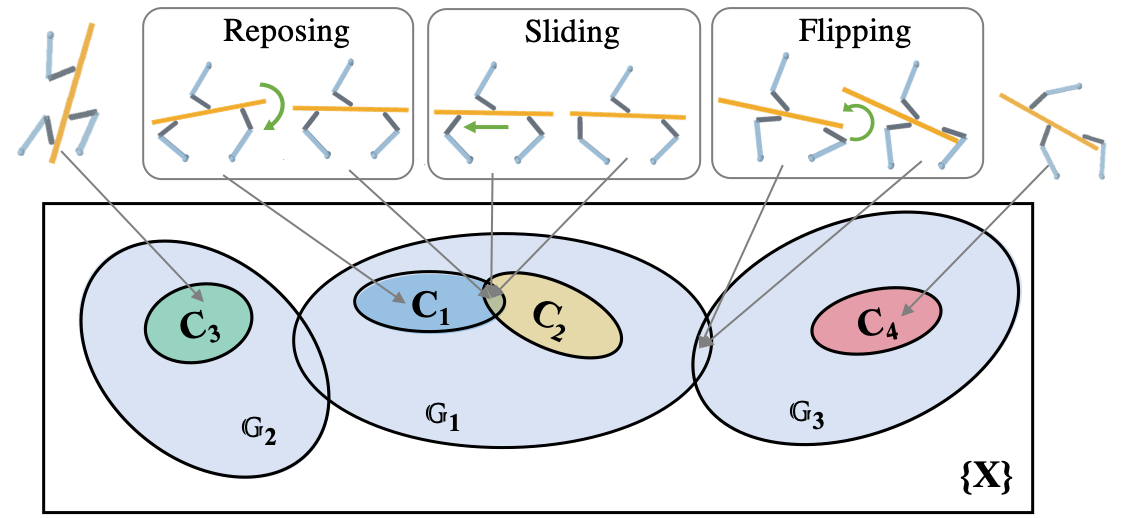}
    \vspace{-9pt}
    \caption{A schematic of the object pose space $\{\boldsymbol{X}\}$, contact configurations $\boldsymbol{C}$, grasp postures $\mathbb{G}$ and motion primitives. Each contact configuration $\boldsymbol{C}$ covers a range of object poses $\boldsymbol{X}$ into which the object can be moved by the hand via \emph{Reposing}. Different $\boldsymbol{C}$s may have some overlapping areas in object pose space such that \emph{Sliding} or \emph{Flipping} can be used to transfer between them. The in-hand manipulation process can be considered as finding a path between two points in $\{\boldsymbol{X}\}$, while different motion primitives are needed to move within or between $\boldsymbol{C}$s.
    } 
    \label{fig:goal_vis}
\end{figure}

\section{Method}
We propose a method to perform in-hand manipulation of objects with a fully-actuated hand, i.e. moving an object from pose $\boldsymbol{X}_0$ to the target pose $\boldsymbol{X}_g$. Specifically, we consider the challenging problem where the object has to be held continuously to be prevented from dropping out of reach. We model the manipulation state with $(\boldsymbol{X, C})$ where $\boldsymbol{X}$ is the object pose relative to the hand, and $\boldsymbol{C}$ is the contact configuration of the fingers relative to the object frame. Our approach is based on the insight that this manipulation state space has a specific structure (see Figure~\ref{fig:goal_vis}). In that space, one object pose can be realized by multiple contact configurations. One contact configuration can move the object to a range of poses. However, some object pose changes may require changing the contact configuration by either sliding on the object or flipping one finger to the other side of the object. Moreover, flipping results in a different grasp posture $\mathbb{G}$ where one finger makes new contact with a different side of the object. 
For this example task, we only assume hard-finger contacts~\cite{salisbury1983kinematic} at the fingertips and precision grasps.  To navigate the structure in this manipulation state space, we design three manipulation primitives that either {\em repose\/} the object without changing the contact configuration, that {\em slide\/} a finger on the object or {\em flip\/} a finger to another side of the object. The last two primitives change the contact configuration and thereby the reachable object poses. 

In this section, we give details on the low-level controllers that implement these three manipulation primitives and on the mid-level policies that learned to select from these primitives to move the object from the current to the goal state. This hierarchical approach allows the robot to navigate the complex manipulation state space and is visualized in Figure~\ref{fig:architecture} (a).
While the overall idea is agnostic to the specific manipulation task, we use pole manipulation as a running example and show later how the idea generalizes to differently shaped objects. Throughout, we assume full observability of the system state and prior knowledge of inertia parameters of the hand and object. In the real-world, the object and hand could be visually tracked~\cite{Cifuentes}. Furthermore, having access to approximate dynamic and geometric models of hand and object is a weak assumption and we show how our method is robust to observation noise and model-inaccuracies.

\subsection{Low-level controller} \label{chpt:LController}
We implement each motion primitive with a low-level, torque controller. Torque control is inherently more robust to uncertainty by providing compliance. This is empirically evidenced by controllers developed for locomotion and multi-contact control \cite{ott2011posture,simpkins2011complex,herzog2016momentum}. 

The low-level controller receives intermediate target object poses $\boldsymbol{X}_d$ or contact configurations $\boldsymbol{C}_d$ from the mid-level controller. By modelling the dynamics of the hand and object, the controllers can compute the necessary joint torques to achieve $\boldsymbol{X}_d$ or $\boldsymbol{C}_d$.
Following~\cite{Prattichizzo2008}, we express the dynamics of the hand when making contact at each fingertip as:
\begin{equation}\label{eq:hand_dynamics}
    \boldsymbol{M}_{hnd}(\boldsymbol{q})\ddot{\boldsymbol{q}}+V(\boldsymbol{q},\dot{\boldsymbol{q}}) + \boldsymbol{J}^T\boldsymbol{\lambda}=\boldsymbol{\tau},
\end{equation}
where $\boldsymbol{M}_{hnd}(\boldsymbol{q})$ is the mass matrix of the hand, $V$ computes the centrifugal and Coriolis terms,  $\boldsymbol{q} \in \mathbb{R}^n$ is a vector containing the $n$ joint angles of the hand, $\boldsymbol{J}$ is the stacked Jacobian for each finger tip, $\boldsymbol{\lambda}$ contains the contact force and moment applied at each contact, and $\boldsymbol{\tau}\in \mathbb{R}^n$ is the torque applied at each joint. Similarly, we express the dynamics of the object as: 
\begin{equation}\label{eq:obj_dynamics}
    \boldsymbol{M}_{obj}(\boldsymbol{X})\dot{\boldsymbol{\nu}}_{obj} - \boldsymbol{G} \boldsymbol{\lambda} = \boldsymbol{g},
\end{equation}
where $\boldsymbol{M}_{obj}(\boldsymbol{X})$ is the mass matrix of the object, $\boldsymbol{\nu}_{obj}$ is the twist of object, $\boldsymbol{G}$ is the grasp matrix, which relates contact forces to a wrench on the object.
$\boldsymbol{g}$ contains the external forces applied on the object, including gravity. 
Then, by plugging Eq.~\ref{eq:obj_dynamics} into  Eq.~\ref{eq:hand_dynamics}, we have
    \begin{equation} \label{eq:reposing_dynamics}
    \boldsymbol{M}_{hnd}(\boldsymbol{q})\ddot{\boldsymbol{q}} + V(\boldsymbol{q},\dot{\boldsymbol{q}}) + 
 \boldsymbol{J}^T \boldsymbol{G}^{-1}(\boldsymbol{M}_{obj}(\boldsymbol{X})\dot{\boldsymbol{\nu}}_{obj}-\boldsymbol{g}) = \boldsymbol{\tau}. 
\end{equation}
Note that this system has multiple solutions, and has to adhere to additional constraints on the contact force. The normal force $\boldsymbol{\lambda}_N$ has to be larger than 0, and the shear force has to be smaller than $\mu \boldsymbol{\lambda}_N$ to prevent slipping, where $\mu$ is the friction coefficient. To address these constraints, we compute the null space of $\boldsymbol{G}$, denoted as $\mathcal{N}(\boldsymbol{G})$, so that for any $\boldsymbol{\lambda_0} \in \mathcal{N}(\boldsymbol{G})$, we have $\boldsymbol{G\lambda_0} = \boldsymbol{0}$. 
We choose some $\boldsymbol{\lambda_0}$ with a sufficient normal force component
such that for $\boldsymbol{\lambda} = \boldsymbol{\lambda}_p + \boldsymbol{\lambda_0}$, where $\boldsymbol{\lambda}_p$ is a particular solution in Eq.~\ref{eq:obj_dynamics}, the sheer force component stays within the friction cone, thus avoiding unintended sliding on the contact surface.
We use $\boldsymbol{\lambda}$ as the target contact force per contact point in Eq.~\ref{eq:obj_dynamics}, which ensures that the hand holds the object firmly during motion and in the presence of some external perturbations.  


\textbf{Reposing:}
The objective of this manipulation primitive is to change the object pose $\boldsymbol{X}$ while maintaining the contact configuration $\boldsymbol{C}$. 
Under this assumption, there is no slip at the fingertips, which means that the twists of the fingertips are the same as of contact points on the objects. 
This yields $ \boldsymbol{J}\dot{\boldsymbol{q}} = \boldsymbol{\nu}_{ftips} =  \boldsymbol{\nu}_{contact} = \boldsymbol{G}^T\boldsymbol{\nu}_{obj}$. 
Considering the definition of the Jacobian ${\boldsymbol{\nu}}_{ftips}=\boldsymbol{J} \dot{\boldsymbol{q}}$, we differentiate it and obtain $\dot{\boldsymbol{\nu}}_{ftips} = \dot{\boldsymbol{J}}\dot{\boldsymbol{q}} + \boldsymbol{J}\ddot{\boldsymbol{q}}$. 
Substituting this into (\ref{eq:reposing_dynamics}), we obtain:
\begin{equation} \label{eq:cartesian_torque}
\begin{split}
    (\boldsymbol{M}_{hnd}(\boldsymbol{q})\boldsymbol{J}^{-1} + \boldsymbol{J}^T\boldsymbol{G}^{-1}\boldsymbol{M}_{obj}\boldsymbol{G}^{-T})\dot{\boldsymbol{\nu}}_{ftips} + \\ (V(\boldsymbol{q},\dot{\boldsymbol{q}})  - \boldsymbol{M}_{hnd}(\boldsymbol{q})\boldsymbol{J}^{-1}\dot{\boldsymbol{J}}\dot{\boldsymbol{q}}) - 
    \boldsymbol{J}^T \boldsymbol{G}^{-1}\boldsymbol{g}  = \boldsymbol{\tau}.
\end{split}
\end{equation}

Given a target object state $(\boldsymbol{X}_d, \dot{\boldsymbol{X}}_d, \ddot{\boldsymbol{X}}_d)$, we can compute a desired fingertip acceleration $\dot{\boldsymbol{\nu}}_{ftips}$ using a PID controller: $\boldsymbol{u}_{ftips}=\ddot{\boldsymbol{X}}_d + \boldsymbol{K}_d(\dot{\boldsymbol{X}}_d-\dot{\boldsymbol{X}}) + \boldsymbol{K}_p (\boldsymbol{X}_d - \boldsymbol{X})$. The error dynamics of this controller are exponentially stable with suitable gain matrices. Substituting $\boldsymbol{u}_{ftips}$ for $\dot{\boldsymbol{\nu}}_{ftips}$ in Eq.~\ref{eq:cartesian_torque} gives the appropriate actuation torque.


\textbf{Sliding: } The objective of this primitive action is to change the contact configuration $\boldsymbol{C}$ by sliding one finger on the object surface while keeping the other finger contacts fixed. Given a target contact configuration $\boldsymbol{C}_d$, we compute the desired joint angles $\boldsymbol{q}_d$ using inverse kinematics. We then use inverse dynamic to compute the joint torque
\begin{equation} \label{eq:inverse_dynamics_joint_space}
    \boldsymbol{\tau} = \boldsymbol{M}_{hnd}(\boldsymbol{q})\boldsymbol{u}_{joint} + V(\boldsymbol{q},\dot{\boldsymbol{q}}) + \boldsymbol{J}^T \boldsymbol{G}^{-1}(\boldsymbol{M}_{obj}(\boldsymbol{X})\dot{\boldsymbol{\nu}}_{obj}-\boldsymbol{g}),
\end{equation}
where the control input is $\boldsymbol{u}_{joint}= - \boldsymbol{K}_d\dot{\boldsymbol{q}} + \boldsymbol{K}_p (\boldsymbol{q}_d - \boldsymbol{q})$.

\textbf{Flipping:} The objective of this primitive action is to move one finger off the object surface to a new contact position while using the other fingers to hold the object. This is also referred to as {\em finger gaiting\/}~\cite{sundaralingam2018geometric} or {\em{pivoting}}~\cite{ma2011dexterity}.
 Compared to the sliding action, this action is more useful when the new contact configuration $\boldsymbol{C}_{t+1}$ is far from the current one $\boldsymbol{C}_t$.
 Specifically, when manipulating a pole-shaped object, we apply the flipping motion when moving a finger from one side of the pole to the other side. The flipping process can be divided into breaking contact while re-balancing the object and making new contact on the moving object. The flipping primitive can be triggered as a whole by the mid-level controller (as shown in Section~\ref{sec:exp_reach} and \ref{sec:robustness}) or broken into two motion primitives for more complex objects and triggered separately (as shown in Section~\ref{sec:cube}). 
 
 


\subsection{Mid-level Controller} \label{chpt:RL}
In our hierarchical structure for in-hand manipulation, we use DRL to learn a policy that can decide the next motion primitive and its parameters, as shown in Figure~\ref{fig:architecture}(a), so that the hand can move the object to the goal pose $\boldsymbol{X}_g$. Given the current state $\boldsymbol{S}$, the policy selects an action $\boldsymbol{A}$ from a discrete set. Each action in that set corresponds to a motion primitive with a specific choice of parameters. Table~\ref{tab:ActionSpace} lists all the possible actions for moving an object in a plane, i.e. changing its 2D position and orientation. Here, the parameters are either the motion direction of the object or contact or which finger to flip. The motion magnitude $\Delta$ is fixed.  
Given $\boldsymbol{A}$, the lower level uses the aforementioned controllers to reach the provided intermediate target object pose or contact configuration. The agent receives a reward when it reaches the final goal.

    
        \begin{table}
        \caption{Discretization of manipulation primitives into 14 possible actions.}
        \makebox[\textwidth][c]
        {\small
            \begin{tabular}{c|c}
                \hline
                Reposing & $+\Delta\boldsymbol{X}^x$, $-\Delta\boldsymbol{X}^x$, $+\Delta\boldsymbol{X}^y$, $-\Delta\boldsymbol{X}^y$, $+\Delta\boldsymbol{X}^\theta$, $-\Delta\boldsymbol{X}^\theta$  \\
                \hline
                Sliding & $+\Delta \boldsymbol{C}^1$, $-\Delta \boldsymbol{C}^1$, $+\Delta \boldsymbol{C}^2$, $-\Delta \boldsymbol{C}^2$, $+\Delta \boldsymbol{C}^3$, $-\Delta \boldsymbol{C}^3$
                \\
                \hline
                Flipping & Flip either left or right finger to the other side \\
                \hline
            \end{tabular}
        }
        \label{tab:ActionSpace}
            \vspace{-8pt}
    \end{table}

    Note that some actions will only be successful in specific contact configurations. For example, flipping one finger can only succeed when the two other fingers are opposing each other, so that they can keep the object in relative balance when the flipping finger is released. Therefore, flipping can be successfully attempted much less frequently than reposing and sliding. Nevertheless, the policy has to discover the conditions for a successful flip through trial and error. To address this, we assign a positive reward specifically for a successful and useful flipping action, and a small negative reward for failures.
    A flipping motion is useful if the new configuration $\boldsymbol{C}$ makes it easier for the object to move towards the goal. Specifically, flipping can be used to give the hand more freedom to re-orient the object. Thus, if after flipping the normal direction of the flipped finger is aligned with the direction of $\boldsymbol{X}^\theta_{g}-\boldsymbol{X}^\theta_t$, we consider the flip to be `useful'.
    The flipping reward also encourages contact configurations that can make a successful flipping motion when necessary.
    Another major cause for failure is losing contact, which commonly happens when the target pose is outside the workspace limits. 
    To filter out those invalid primitive choices, we design a feasibility filter that rejects actions at states that are not feasible according to inverse kinematics, which also returns a negative reward to the DRL network. This filter is optional, but useful for pruning actions from the policy's output without changing the environment and forcing the policy to pick feasible actions. This approach is similar to training a policy restricted to legal actions, as done in DQN~\cite{mnih2013playing} or AlphaGo~\cite{silver2016mastering}. 
    It is not straightforward to include such a filter in an end-to-end framework that directly outputs joint torques without a notion of workspace limits.
    
\section{Experiments}
\label{chpt:experiment}
        In this section, we report experimental results regarding the following questions: (i) How well does our approach perform in-hand manipulation tasks
        ? (ii) How robust is our model in the presence of observation noise and model inaccuracies on, for instance, object dimensions and weight? And (iii) does our approach generalize to objects of different shapes? We compare to an end-to-end and search-based baseline. Our hypothesis is that our approach is especially advantageous for object goal poses that require complex manipulation sequences.
    
    We implemented our approach in PyBullet~\cite{coumans2019} in which a three-fingered hand has to move an object within a 2D vertical plane from one pose to a goal pose. Downward gravity is present. Therefore, the object has 3DoF (2D position and 1D orientation) and has to be held during manipulation to not drop. Each finger is fully-actuated and has 2DoF. For the first two experiments, we use a pole-shaped object of $50$cm length and $2$cm width. For the third experiment, we use a cube.

\subsection{Details on Training Mid-Level Policies and Baseline}
    We train the proposed mid-level  policy 
    using {\em Proximal Policy Optimization\/} (PPO)~\cite{schulman2017proximal} with 16 parallel environments. The observation includes joint angles $\boldsymbol{q}$ and velocities $\dot{\boldsymbol{q}}$, current object pose $\boldsymbol{X}_t$ and the goal pose $\boldsymbol{X}_g$. The reward function is as follows
    \begin{equation*}    
      r_t = \begin{cases}
      +5 &    \text{if the object reaches the goal at time $t$,} \\
      -0.01 & \text{if the action is invalid at time $t$,} \\
      +1 &    \text{if flipping is useful and successful.}
      \end{cases}
    \end{equation*}
    An episode is considered successful if the object reaches a pose close to the goal, with a translational error less than $1$cm and rotational error less than $5.7^\circ$ ($0.1$ rad). Since the hand should always firmly grasp the object, we consider dropping the object a failure case. This happens when there is no contact on one of the sides of the object. It is also a failure if the hand does not reach the goal within a time limit. 
    
    We test the effectiveness and robustness of our method in comparison to an end-to-end policy trained using {\em Deep Deterministic Policy Gradient\/} (DDPG)~\cite{lillicrap2015continuous}. This policy operates in the same simulation environment and directly outputs joint torques without encoding any structural priors in the form of traditional controllers or the dynamics of the system. We chose DDPG as it has been shown to work well for continuous control tasks, and for complex dexterous manipulation~\cite{rajeswaran2017learning,Zhu2018DexterousMW}. The reward function is sparse with $+5$ when reaching the goal otherwise $0$. Both policies are represented by $3$ fully connected layers with $256$ units.
    
    We also compare to a search-based baseline~\cite{cruciani2019dual}. For this, we build a graph by discretizing the feasible object poses $\boldsymbol{X}$ and contact configurations $\boldsymbol{C}$. To trade-off computational complexity and precision, we empirically choose a grid resolution of $2$cm and $0.2$rad for $\boldsymbol{X}$ and $5$cm for $\boldsymbol{C}$. The grid cells form the nodes in the graph. The edges between neighboring nodes correspond to the manipulation primitive that connects them. In-hand manipulation is then equivalent to finding a path from the current pose to the goal pose using Dijkstra's algorithm~\cite{dijkstra1959note}.
    
\subsection{Dataset}
To train our model on reachable goals, we collect a comprehensive dataset of the object's initial poses $\boldsymbol{X}_0$ and goal poses $\boldsymbol{X}_g$. 
We divide the dataset of $(\boldsymbol{X}_0, \boldsymbol{X}_g)$ pairs into three groups: Easy, Medium, and Hard Goals. In the Easy Goal group, $\boldsymbol{X}_0$ and $\boldsymbol{X}_g$ are in the range of the same contact configuration $\boldsymbol{C}$. Therefore, the goal can be reached by only using reposing. The Medium Goal group contains pairs $\boldsymbol{X}_0$ and $\boldsymbol{X}_g$ that are within the same grasp posture $\mathbb{G}$. Only sliding and reposing may be required to reach the goal pose. The Hard Goal group includes pairs of arbitrary $(\boldsymbol{X}_0, \boldsymbol{X}_g)$ pairs, some of which require flipping to be successful.

\begin{table}
        \centering
\small
        \caption{Success rate (with standard deviation) for 500 episodes of manipulating a pole averaged over 3 random seeds. Dropping is a common failure case when the hand loses necessary contact to keep the object grasped. Average time represents the real-world time to complete a successful manipulation sequence. Experiments are conducted in three groups with an increasing level of difficulty.
        }
    \makebox[\textwidth][c]
    {
        \begin{tabular}{l|cccccccccc}
            \hline
              & Succ. Rate &  Drop Rate & Avg. Time (s) \\
            \hline
            Ours-Easy&95.4\%$\pm$0.7\%&3.8\%$\pm$0.7\%&7.1$\pm$0.2\\
            Ours-Mid & 94.6\%$\pm$0.4\%&3.7\%$\pm$1.3\%&8.9$\pm$0.6\\
            Ours-Hard  & 79.3\%$\pm$0.6\%&15.7\%$\pm$1.4\%&13.4$\pm$0.3\\
            \hline
            DDPG-Easy & 72.5\%$\pm$13.5\%  & 19.3\%$\pm$14.2\%& 1.82$\pm$0.1 \\
            DDPG-Mid & 61.7\%$\pm$6.6\% & 26.8\%$\pm$10.2\% &1.97$\pm$0.1 \\
            DDPG-Hard & 11.1\%$\pm$6.8\%& 84.0\%$\pm$10.8\%& 2.35$\pm$0.1  \\
            \hline
            Search-Easy & 91.4\% & 5.2\% & 8.1 \\
            Search-Mid & 79.8\% & 13.6\% & 8.1 \\
            Search-Hard & 54.4\% & 34.8\% & 8.25\\
            \hline
        \end{tabular}
    }
    \label{tab:comparison}
    \end{table}
    
\subsection{Reaching Desired Object Poses}
\label{chpt:Experiment1}
We evaluate the performance of our method on the task of moving the object to desired goal poses and compare to an end-to-end and search-based baseline. 
%
%
We report results in Table~\ref{tab:comparison}. Our method has a consistently higher success and lower dropping rate. The gap to the baselines is especially high for the hard goals. In fact, the end-to-end method fails almost all cases when flipping is required. The results indicate that our hierarchical method has significant advantages when a complex manipulation sequence is required. 

An advantage of the end-to-end method is that the resulting policies yield much faster manipulation sequences as can be seen in the supplementary video. The agent learns to exploit the fast dynamics of the object and frequently pushes it so that it is flying towards the goal pose. However, these dynamic manipulation behaviors will hardly be robust when implemented on a real robotic hand and just as in simulation will likely result in the object being dropped frequently.  
Since our method uses pre-defined motion primitives for manipulation, it does not learn these fast behaviors. However, it will drop the object less often. In the next section, we also show that the overall approach is more robust as the traditional, compliant torque controllers can cope with some observation noise, external perturbations and modelling errors. We expect this robustness to translate to the real world as well. 
Another advantage of the end-to-end method is that it can flexibly choose the contact points on the hand and exploit other links of the fingers for manipulation than just the tips. While our method does this to some extend, we aim to explicitly add more manipulation primitives and possible contact points on the finger links.

Also the search-based baseline cannot achieve as high of a success rate and as low of a drop rate compared to our approach. A major disadvantage of the search-based method is that its computational complexity grows in $O(n^6)$ 
with an increasing resolution of $\boldsymbol{X}$ and $\boldsymbol{C}$. For example, the number of nodes can grow to $10^{9}$ with a $1$cm position resolution and $0.1$rad angle resolution for the medium goal tasks. However, a lower resolution results in a lower accuracy and therefore in a lower planning performance. For example, the flipping primitive is hard to put in a low-resolution graph because this primitive is conditioned on the two fingers that remain in contact oppose each other. 

One limitation of our method is the fixed step size of the manipulation sequence. It may be too small in the early stages of motion, and too large when being already close to the goal. In future work, we will explore continuous mid-level controllers or adaptive step sizes.

\label{sec:exp_reach}
    

    \begin{figure}
        \centering
        \small
        \includegraphics[width = \textwidth]{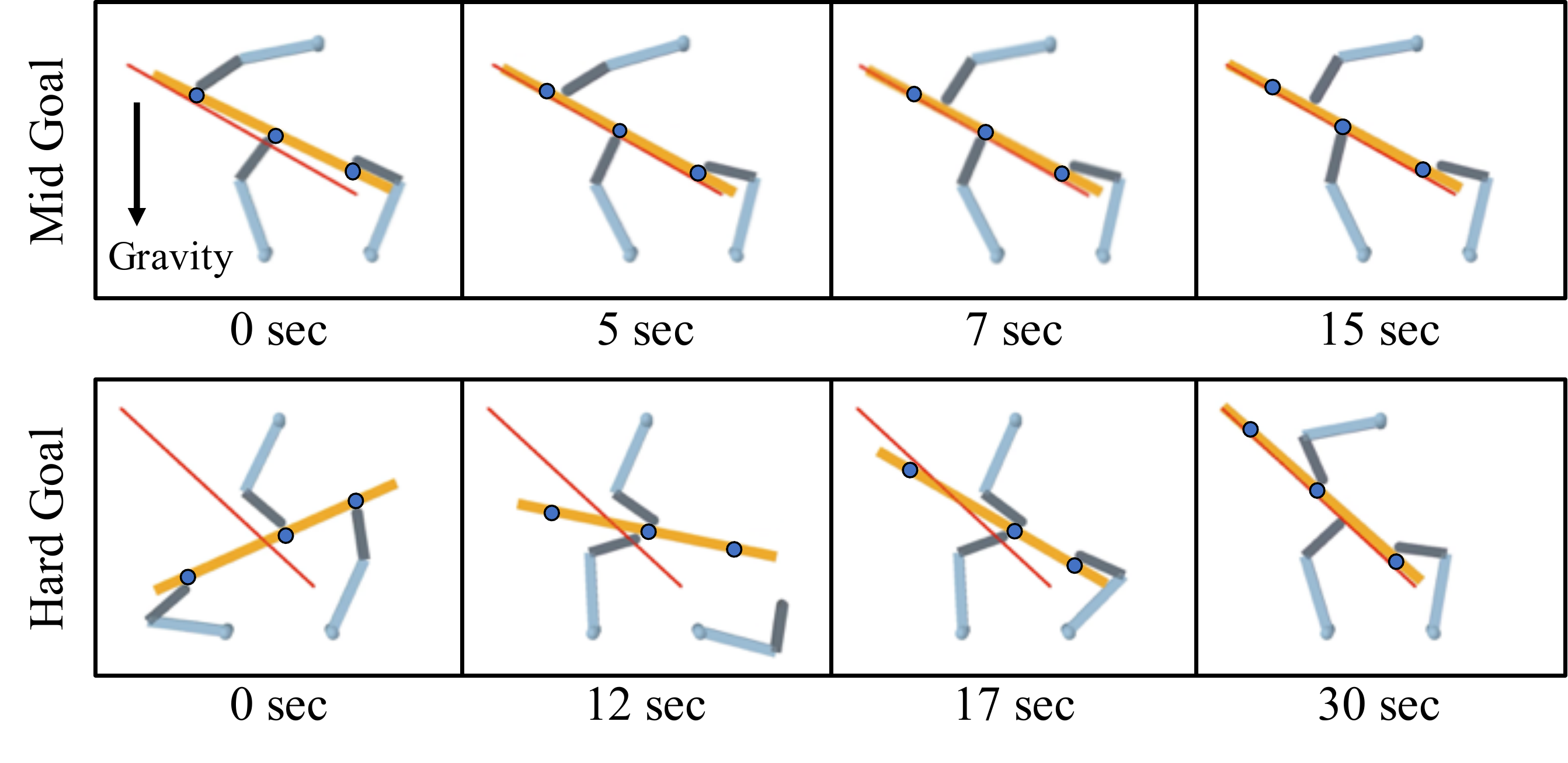}
        \vspace{-7pt}
        \caption{Snapshots of some example episodes when manipulating the pole to the goal pose. The goal pose is indicated in red and the initial contact points are indicated in blue. The upper episode shows the medium goal and the lower episode shows the hard goal.}
        \label{fig:EpisodeExample}
        \vspace{-10pt}
    \end{figure}

    \begin{table*}
    \small
    \caption{Success rate for inaccurate object models (geometry and dynamics) and/or observation noise. Experiments are done with the medium goal group. The first row shows results for the model that is trained and tested without observation noise. The second row reports results for the model that is trained and tested with observation noise.}
        \begin{tabular}{c|ccccccccc}
            \hline
              Training \& Testing & Trained Obj. & Heavier & Lighter & Thicker & Thinner & Longer & Shorter \\ \hline
             Without noise $\epsilon_{x,y,\theta}$ & 94.4\%& 94.4\%& 95.8\%& 91.8\%& 94.2\%& 91.8\% & 95.2\%\\ 
             With noise $\epsilon_{x,y,\theta}$ & 93.2\%& 92.4\%& 94.6\%& 94.4\%& 91\%& 89.2\%&92.6\%\\
            \hline
        \end{tabular} 
        \label{tab:robust}
    \end{table*}

\subsection{Robustness Analysis}
\label{sec:robustness}

In this section, we evaluate the robustness of our method in the presence of observation noise and inaccurate models of the object's geometry and inertia properties. Different from end-to-end models (which typically train one policy per object), traditional controllers require some prior knowledge on the robot hand and object. In the real world, object pose estimation may be noisy and the object and hand models may be inaccurate. A successful manipulation method should be robust to these real-world effects. 

    We evaluate the robustness of our method by adding observation noise to the object pose $\boldsymbol{X}_t$ and by changing the geometry and inertia parameters of the object from the training conditions.
    We designed the dynamic controller and trained the DRL mid-level controller as described in Section~\ref{sec:exp_reach} (Trained Obj.), and test it on poles that are heavier or lighter (1.1x/0.9x mass), thicker or thinner (1.5x/0.5x width), and longer or shorter (1.1x/0.9x length). 
    In the second experiment, we add uniformly distributed observation noise to the object pose $\boldsymbol{X}_t$ with error bounds of $[-0.5,0.5]$ (cm) in translation and $[-2.86, 2.86]$ (deg) in rotation. In Table~\ref{tab:robust}, we report the success rate of 500 evaluation trials on each of these perturbations.
    The result shows that our method is very robust to the inaccuracy of the object model as the performance is almost the same. When there is observation noise in the system, the success rate drops a few percentage points but the system still performs well. Observation noise leads to error in the computed contact configurations and thereby adds noise to the low-level dynamic controllers. However, two factors make our approach robust to this. First, the mid-level controller takes state feedback into account at every step and is therefore able to react to unexpected changes. 
    Second, torque control is inherently robust to uncertainty due to compliance. For some cases when one finger makes pre-mature contact, the torque will be achieved earlier than expected and the finger will compliantly give in.

\subsection{Manipulating a cube}
\label{sec:cube}
\begin{figure}
    \centering
    \small
    \includegraphics[width = 0.95\linewidth]{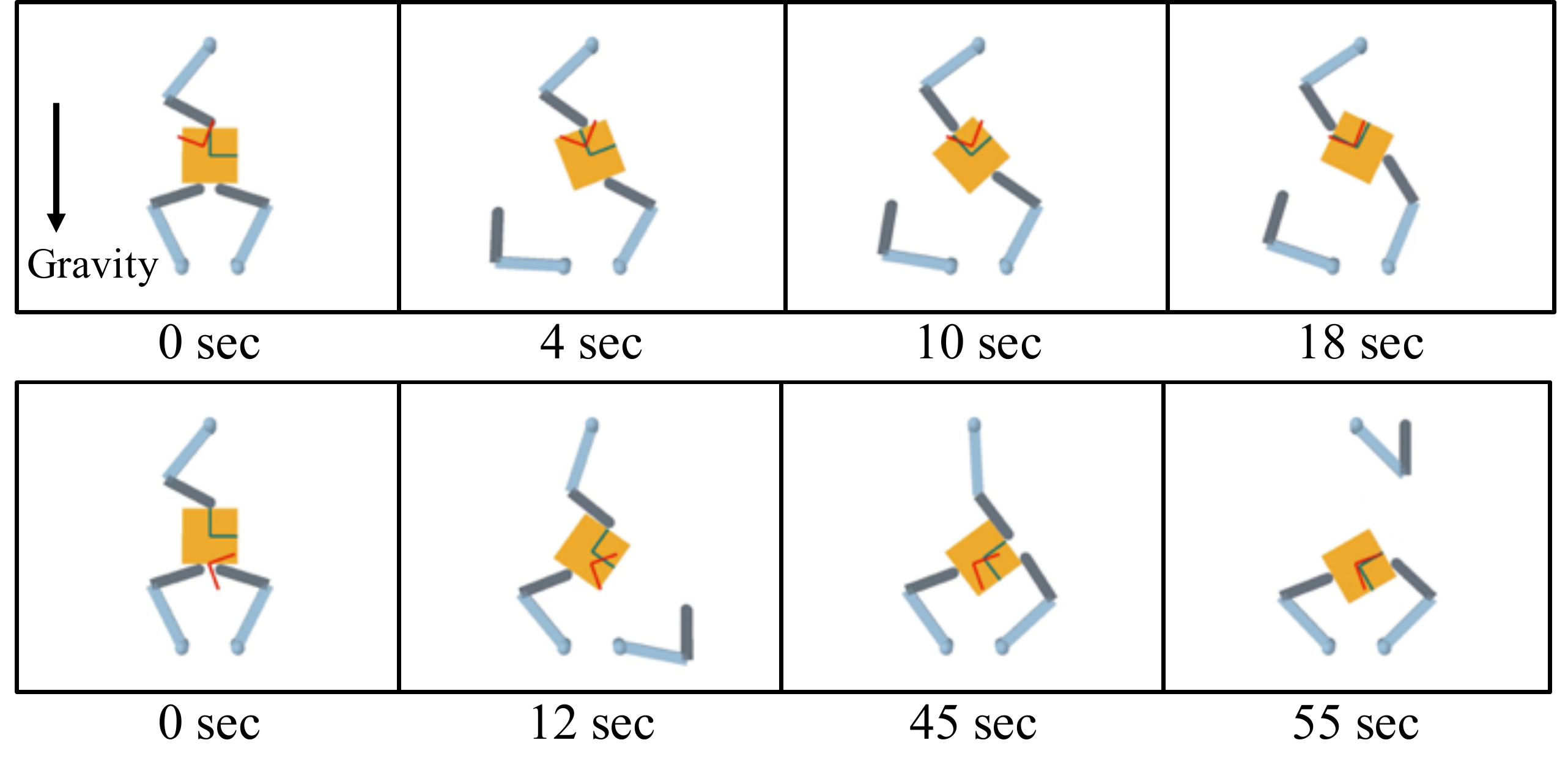}
    \vspace{-5pt}
    \caption{Examples for manipulating a small cube. The goal pose is indicated in red.
    }
    \label{fig:CubeExperiment}
    \vspace{-10pt}
\end{figure}

Lastly, we test whether the overall approach of combining learned policies with the three available manipulation primitives generalizes to objects that are not pole-shaped. We consider a cube (Fig.~\ref{fig:CubeExperiment}) as an example. Compared to poles, the cube is shorter (side length of $10$cm) and has 4 edges instead of 2, and therefore has more grasp postures $\mathbb{G}$.
Ideally, the mid-level controller should learn how to choose the motion primitives and parameters, especially for flipping. But for the sake of faster training for this proof of concept with a small object, we disable sliding since it has a limited benefit on expanding the range of object poses. We also decompose flipping into two stages: releasing and landing. 
In the experiment, we choose one initial object pose $\boldsymbol{X}_0$, and sample any reachable pose $\boldsymbol{X}$ in the range as the goal pose. On a test set of $500$ episodes, our method achieves a success rate of $71.4\%$, and a dropping rate of $20.8\%$. The average running time is $7.8$s. The results show that our overall approach can be generalized to objects of varying shapes. When inspecting the manipulation behaviour (see supplementary video), we find that releasing is often used because it is easy to balance a short object with two fingers. The mid-level controller learned to use the two-fingered primitives to manipulate the cube in most of the cases, because those two finger configurations cover a larger range of object poses $\boldsymbol{X}$, and they can balance the small cube well. 

\section{Conclusion}
\label{sec:conclusion}

We proposed a hierarchical approach to robotic in-hand manipulation, and evaluated our method in simulation with a 3-fingered hand that moves a pole-shaped object in a plane. For the low-level control, our method uses model-based, torque controllers that require some prior knowledge on geometry and inertia of hand and object. They control a set of manipulation primitives. We proposed to learn a mid-level controller with DRL to generate a sequence of these primitives for reaching a distant object goal pose. Using this hierarchical architecture, that it significantly outperformed an end-to-end and search-based baseline while also dropping the object less often. In particular, our approach is better for more distant object goal poses that require complex manipulation sequences.
We also showed that our method is robust to observation noise and inaccuracies of the object models in geometry and inertia properties. We also showed how the overall approach can be generalized to objects with different shapes. As future work, we aim to extend the number of manipulation primitives and possible contact points. We also aim to test the approach on a dexterous hand, first in simulation and then in the real world.



\clearpage
\bibliography{ref_inhand2D}

\begin{thebibliography}{41}
\providecommand{\natexlab}[1]{#1}
\providecommand{\url}[1]{#1}
\csname url@samestyle\endcsname
\providecommand{\newblock}{\relax}
\providecommand{\bibinfo}[2]{#2}
\providecommand{\BIBentrySTDinterwordspacing}{\spaceskip=0pt\relax}
\providecommand{\BIBentryALTinterwordstretchfactor}{4}
\providecommand{\BIBentryALTinterwordspacing}{\spaceskip=\fontdimen2\font plus
\BIBentryALTinterwordstretchfactor\fontdimen3\font minus
  \fontdimen4\font\relax}
\providecommand{\BIBforeignlanguage}[2]{{%
\expandafter\ifx\csname l@#1\endcsname\relax
\typeout{** WARNING: IEEEtranN.bst: No hyphenation pattern has been}%
\typeout{** loaded for the language `#1'. Using the pattern for}%
\typeout{** the default language instead.}%
\else
\language=\csname l@#1\endcsname
\fi
#2}}
\providecommand{\BIBdecl}{\relax}
\BIBdecl

\bibitem[Okamura et~al.(2000)Okamura, Smaby, and Cutkosky]{okamura2000overview}
A.~M. Okamura, N.~Smaby, and M.~R. Cutkosky, ``An overview of dexterous
  manipulation,'' in \emph{Proceedings 2000 ICRA. Millennium Conference. IEEE
  International Conference on Robotics and Automation. Symposia Proceedings
  (Cat. No. 00CH37065)}, vol.~1.\hskip 1em plus 0.5em minus 0.4em\relax IEEE,
  2000, pp. 255--262.

\bibitem[{Bicchi}(2000)]{Bicchi2000}
A.~{Bicchi}, ``Hands for dexterous manipulation and robust grasping: a
  difficult road toward simplicity,'' \emph{IEEE Transactions on Robotics and
  Automation}, vol.~16, no.~6, pp. 652--662, Dec 2000.

\bibitem[Ma and Dollar(2011)]{ma2011dexterity}
R.~R. Ma and A.~M. Dollar, ``On dexterity and dexterous manipulation,'' in
  \emph{2011 15th International Conference on Advanced Robotics (ICAR)}.\hskip
  1em plus 0.5em minus 0.4em\relax IEEE, 2011, pp. 1--7.

\bibitem[Jacobsen et~al.(1986)Jacobsen, Iversen, Knutti, Johnson, and
  Biggers]{MITUtahHand}
S.~Jacobsen, E.~Iversen, D.~Knutti, R.~Johnson, and K.~Biggers, ``Design of the
  utah/mit dextrous hand,'' in \emph{Proceedings. 1986 IEEE International
  Conference on Robotics and Automation}, vol.~3.\hskip 1em plus 0.5em minus
  0.4em\relax IEEE, 1986, pp. 1520--1532.

\bibitem[Kawasaki et~al.(2002)Kawasaki, Komatsu, and Uchiyama]{GifuHand}
H.~Kawasaki, T.~Komatsu, and K.~Uchiyama, ``Dexterous anthropomorphic robot
  hand with distributed tactile sensor: Gifu hand ii,'' \emph{IEEE/ASME
  transactions on mechatronics}, vol.~7, no.~3, pp. 296--303, 2002.

\bibitem[Lee et~al.(2016)Lee, Park, Park, Baeg, and Bae]{AllegroHand}
D.-H. Lee, J.-H. Park, S.-W. Park, M.-H. Baeg, and J.-H. Bae, ``Kitech-hand: A
  highly dexterous and modularized robotic hand,'' \emph{IEEE/ASME Transactions
  on Mechatronics}, vol.~22, no.~2, pp. 876--887, 2016.

\bibitem[Rus(1992)]{rus1992dexterous}
D.~Rus, ``Dexterous rotations of polyhedra,'' in \emph{Proceedings 1992 IEEE
  International Conference on Robotics and Automation}.\hskip 1em plus 0.5em
  minus 0.4em\relax IEEE, 1992, pp. 2758--2763.

\bibitem[Li et~al.(2013)Li, Elbrechter, Haschke, and Ritter]{li2013integrating}
Q.~Li, C.~Elbrechter, R.~Haschke, and H.~Ritter, ``Integrating vision, haptics
  and proprioception into a feedback controller for in-hand manipulation of
  unknown objects,'' in \emph{2013 IEEE/RSJ International Conference on
  Intelligent Robots and Systems}.\hskip 1em plus 0.5em minus 0.4em\relax IEEE,
  2013, pp. 2466--2471.

\bibitem[Rus(1997)]{rus1997coordinated}
D.~Rus, ``Coordinated manipulation of objects in a plane,''
  \emph{Algorithmica}, vol.~19, no. 1-2, pp. 129--147, 1997.

\bibitem[Sundaralingam and Hermans(2017)]{sundaralingam2017relaxed}
B.~Sundaralingam and T.~Hermans, ``Relaxed-rigidity constraints: In-grasp
  manipulation using purely kinematic trajectory optimization,''
  \emph{planning}, vol.~6, p.~7, 2017.

\bibitem[Calli and Dollar(2016)]{calli2016vision}
B.~Calli and A.~M. Dollar, ``Vision-based precision manipulation with
  underactuated hands: Simple and effective solutions for dexterity,'' in
  \emph{2016 IEEE/RSJ International Conference on Intelligent Robots and
  Systems (IROS)}.\hskip 1em plus 0.5em minus 0.4em\relax IEEE, 2016, pp.
  1012--1018.

\bibitem[Kumar et~al.(2016)Kumar, Todorov, and Levine]{kumar2016optimal}
V.~Kumar, E.~Todorov, and S.~Levine, ``Optimal control with learned local
  models: Application to dexterous manipulation,'' in \emph{2016 IEEE
  International Conference on Robotics and Automation (ICRA)}.\hskip 1em plus
  0.5em minus 0.4em\relax IEEE, 2016, pp. 378--383.

\bibitem[Andrychowicz et~al.(2018)Andrychowicz, Baker, Chociej, Jozefowicz,
  McGrew, Pachocki, Petron, Plappert, Powell, Ray, et~al.]{OpenAIHand}
M.~Andrychowicz, B.~Baker, M.~Chociej, R.~Jozefowicz, B.~McGrew, J.~Pachocki,
  A.~Petron, M.~Plappert, G.~Powell, A.~Ray \emph{et~al.}, ``Learning dexterous
  in-hand manipulation,'' \emph{arXiv preprint arXiv:1808.00177}, 2018.

\bibitem[Zhu et~al.(2019)Zhu, Gupta, Rajeswaran, Levine, and
  Kumar]{Zhu2018DexterousMW}
H.~Zhu, A.~Gupta, A.~Rajeswaran, S.~Levine, and V.~Kumar, ``Dexterous
  manipulation with deep reinforcement learning: Efficient, general, and
  low-cost,'' in \emph{2019 International Conference on Robotics and Automation
  (ICRA)}.\hskip 1em plus 0.5em minus 0.4em\relax IEEE, 2019, pp. 3651--3657.

\bibitem[Sundaralingam and Hermans(2018)]{sundaralingam2018geometric}
B.~Sundaralingam and T.~Hermans, ``Geometric in-hand regrasp planning:
  Alternating optimization of finger gaits and in-grasp manipulation,'' in
  \emph{2018 IEEE International Conference on Robotics and Automation
  (ICRA)}.\hskip 1em plus 0.5em minus 0.4em\relax IEEE, 2018, pp. 231--238.

\bibitem[Cruciani et~al.(2019)Cruciani, Hang, Smith, and
  Kragic]{cruciani2019dual}
S.~Cruciani, K.~Hang, C.~Smith, and D.~Kragic, ``Dual-arm in-hand manipulation
  and regrasping using dexterous manipulation graphs,'' \emph{arXiv preprint
  arXiv:1904.11382}, 2019.

\bibitem[Okada(1982)]{okada1982computer}
T.~Okada, ``Computer control of multijointed finger system for precise
  object-handling,'' \emph{IEEE Transactions on Systems, Man, and Cybernetics},
  vol.~12, no.~3, pp. 289--299, 1982.

\bibitem[Kerr and Roth(1986)]{kerr1986analysis}
J.~Kerr and B.~Roth, ``Analysis of multifingered hands,'' \emph{The
  International Journal of Robotics Research}, vol.~4, no.~4, pp. 3--17, 1986.

\bibitem[Leveroni and Salisbury(1996)]{leveroni1996reorienting}
S.~Leveroni and K.~Salisbury, ``Reorienting objects with a robot hand using
  grasp gaits,'' in \emph{Robotics Research}.\hskip 1em plus 0.5em minus
  0.4em\relax Springer, 1996, pp. 39--51.

\bibitem[Mordatch et~al.(2012)Mordatch, Popovi{\'c}, and
  Todorov]{mordatch2012contact}
I.~Mordatch, Z.~Popovi{\'c}, and E.~Todorov, ``Contact-invariant optimization
  for hand manipulation,'' in \emph{Proceedings of the ACM
  SIGGRAPH/Eurographics symposium on computer animation}.\hskip 1em plus 0.5em
  minus 0.4em\relax Eurographics Association, 2012, pp. 137--144.

\bibitem[Fan et~al.(2017)Fan, Tang, Lin, Zhao, and Tomizuka]{fan2017real}
Y.~Fan, T.~Tang, H.-C. Lin, Y.~Zhao, and M.~Tomizuka, ``Real-time robust finger
  gaits planning under object shape and dynamics uncertainties,'' in \emph{2017
  IEEE/RSJ International Conference on Intelligent Robots and Systems
  (IROS)}.\hskip 1em plus 0.5em minus 0.4em\relax IEEE, 2017, pp. 1267--1273.

\bibitem[Odhner and Dollar(2015)]{odhner2015stable}
L.~U. Odhner and A.~M. Dollar, ``Stable, open-loop precision manipulation with
  underactuated hands,'' \emph{The International Journal of Robotics Research},
  vol.~34, no.~11, pp. 1347--1360, 2015.

\bibitem[Calli and Dollar(2018)]{calli2018robust}
B.~Calli and A.~M. Dollar, ``Robust precision manipulation with simple process
  models using visual servoing techniques with disturbance rejection,''
  \emph{IEEE Transactions on Automation Science and Engineering}, no.~99, pp.
  1--14, 2018.

\bibitem[Platt et~al.(2004)Platt, Fagg, and Grupen]{platt2004manipulation}
R.~Platt, A.~H. Fagg, and R.~A. Grupen, ``Manipulation gaits: Sequences of
  grasp control tasks,'' in \emph{IEEE International Conference on Robotics and
  Automation, 2004. Proceedings. ICRA'04. 2004}, vol.~1.\hskip 1em plus 0.5em
  minus 0.4em\relax IEEE, 2004, pp. 801--806.

\bibitem[Simpkins and Todorov(2011)]{simpkins2011complex}
A.~Simpkins and E.~Todorov, ``Complex object manipulation with hierarchical
  optimal control,'' in \emph{2011 IEEE Symposium on Adaptive Dynamic
  Programming and Reinforcement Learning (ADPRL)}.\hskip 1em plus 0.5em minus
  0.4em\relax IEEE, 2011, pp. 338--345.

\bibitem[Murooka et~al.(2017)Murooka, Ueda, Nozawa, Kakiuchi, Okada, and
  Inaba]{murooka2017global}
M.~Murooka, R.~Ueda, S.~Nozawa, Y.~Kakiuchi, K.~Okada, and M.~Inaba, ``Global
  planning of whole-body manipulation by humanoid robot based on transition
  graph of object motion and contact switching,'' \emph{Advanced Robotics},
  vol.~31, no.~6, pp. 322--340, 2017.

\bibitem[Van~Hoof et~al.(2015)Van~Hoof, Hermans, Neumann, and
  Peters]{van2015learning}
H.~Van~Hoof, T.~Hermans, G.~Neumann, and J.~Peters, ``Learning robot in-hand
  manipulation with tactile features,'' in \emph{2015 IEEE-RAS 15th
  International Conference on Humanoid Robots (Humanoids)}.\hskip 1em plus
  0.5em minus 0.4em\relax IEEE, 2015, pp. 121--127.

\bibitem[Rajeswaran* et~al.(2018)Rajeswaran*, Kumar*, Gupta, Vezzani, Schulman,
  Todorov, and Levine]{rajeswaran2017learning}
A.~Rajeswaran*, V.~Kumar*, A.~Gupta, G.~Vezzani, J.~Schulman, E.~Todorov, and
  S.~Levine, ``{Learning Complex Dexterous Manipulation with Deep Reinforcement
  Learning and Demonstrations},'' in \emph{Proceedings of Robotics: Science and
  Systems (RSS)}, 2018.

\bibitem[Silver et~al.(2018)Silver, Allen, Tenenbaum, and
  Kaelbling]{silver2018residual}
T.~Silver, K.~Allen, J.~Tenenbaum, and L.~Kaelbling, ``Residual policy
  learning,'' \emph{arXiv preprint arXiv:1812.06298}, 2018.

\bibitem[Johannink et~al.(2019)Johannink, Bahl, Nair, Luo, Kumar, Loskyll,
  Ojea, Solowjow, and Levine]{johannink2018residual}
T.~Johannink, S.~Bahl, A.~Nair, J.~Luo, A.~Kumar, M.~Loskyll, J.~A. Ojea,
  E.~Solowjow, and S.~Levine, ``Residual reinforcement learning for robot
  control,'' in \emph{2019 International Conference on Robotics and Automation
  (ICRA)}.\hskip 1em plus 0.5em minus 0.4em\relax IEEE, 2019, pp. 6023--6029.

\bibitem[Salisbury and Roth(1983)]{salisbury1983kinematic}
J.~K. Salisbury and B.~Roth, ``Kinematic and force analysis of articulated
  mechanical hands,'' \emph{Journal of Mechanisms, Transmissions, and
  Automation in Design}, vol. 105, no.~1, pp. 35--41, 1983.

\bibitem[{Garcia Cifuentes} et~al.(2017){Garcia Cifuentes}, {Issac},
  {Wüthrich}, {Schaal}, and {Bohg}]{Cifuentes}
C.~{Garcia Cifuentes}, J.~{Issac}, M.~{Wüthrich}, S.~{Schaal}, and J.~{Bohg},
  ``Probabilistic articulated real-time tracking for robot manipulation,''
  \emph{IEEE Robotics and Automation Letters}, vol.~2, no.~2, pp. 577--584,
  April 2017.

\bibitem[Ott et~al.(2011)Ott, Roa, and Hirzinger]{ott2011posture}
C.~Ott, M.~A. Roa, and G.~Hirzinger, ``Posture and balance control for biped
  robots based on contact force optimization,'' in \emph{2011 11th IEEE-RAS
  International Conference on Humanoid Robots}.\hskip 1em plus 0.5em minus
  0.4em\relax IEEE, 2011, pp. 26--33.

\bibitem[Herzog et~al.(2016)Herzog, Rotella, Mason, Grimminger, Schaal, and
  Righetti]{herzog2016momentum}
A.~Herzog, N.~Rotella, S.~Mason, F.~Grimminger, S.~Schaal, and L.~Righetti,
  ``Momentum control with hierarchical inverse dynamics on a torque-controlled
  humanoid,'' \emph{Autonomous Robots}, vol.~40, no.~3, pp. 473--491, 2016.

\bibitem[Prattichizzo and Trinkle(2008)]{Prattichizzo2008}
D.~Prattichizzo and J.~C. Trinkle, \emph{Grasping}.\hskip 1em plus 0.5em minus
  0.4em\relax Berlin, Heidelberg: Springer Berlin Heidelberg, 2008, pp.
  671--700.

\bibitem[Mnih et~al.(2013)Mnih, Kavukcuoglu, Silver, Graves, Antonoglou,
  Wierstra, and Riedmiller]{mnih2013playing}
V.~Mnih, K.~Kavukcuoglu, D.~Silver, A.~Graves, I.~Antonoglou, D.~Wierstra, and
  M.~Riedmiller, ``Playing atari with deep reinforcement learning,''
  \emph{arXiv preprint arXiv:1312.5602}, 2013.

\bibitem[Silver et~al.(2016)Silver, Huang, Maddison, Guez, Sifre, Van
  Den~Driessche, Schrittwieser, Antonoglou, Panneershelvam, Lanctot,
  et~al.]{silver2016mastering}
D.~Silver, A.~Huang, C.~J. Maddison, A.~Guez, L.~Sifre, G.~Van Den~Driessche,
  J.~Schrittwieser, I.~Antonoglou, V.~Panneershelvam, M.~Lanctot \emph{et~al.},
  ``Mastering the game of go with deep neural networks and tree search,''
  \emph{nature}, vol. 529, no. 7587, p. 484, 2016.

\bibitem[Coumans and Bai(2016--2019)]{coumans2019}
E.~Coumans and Y.~Bai, ``Pybullet, a python module for physics simulation for
  games, robotics and machine learning,'' \url{http://pybullet.org},
  2016--2019.

\bibitem[Schulman et~al.(2017)Schulman, Wolski, Dhariwal, Radford, and
  Klimov]{schulman2017proximal}
J.~Schulman, F.~Wolski, P.~Dhariwal, A.~Radford, and O.~Klimov, ``Proximal
  policy optimization algorithms,'' \emph{arXiv preprint arXiv:1707.06347},
  2017.

\bibitem[Lillicrap et~al.(2015)Lillicrap, Hunt, Pritzel, Heess, Erez, Tassa,
  Silver, and Wierstra]{lillicrap2015continuous}
T.~P. Lillicrap, J.~J. Hunt, A.~Pritzel, N.~Heess, T.~Erez, Y.~Tassa,
  D.~Silver, and D.~Wierstra, ``Continuous control with deep reinforcement
  learning,'' \emph{arXiv preprint arXiv:1509.02971}, 2015.

\bibitem[Dijkstra(1959)]{dijkstra1959note}
E.~W. Dijkstra, ``A note on two problems in connexion with graphs,''
  \emph{Numerische mathematik}, vol.~1, no.~1, pp. 269--271, 1959.

\end{thebibliography}
\bibliographystyle{IEEEtranN}
\clearpage

\end{document}